\documentclass[a4paper, 10 pt, conference]{ieeeconf} 

\IEEEoverridecommandlockouts                             
\usepackage{algorithm}
\usepackage{algorithmic}
\usepackage{amsmath}
\usepackage{graphicx}
\usepackage{textcomp}
\graphicspath{ {Images/} }

\title{\LARGE \bf
A Modular Spatial Clustering Algorithm with Noise
Specification}

\author{ \parbox{3.3 in}{\centering Akhil K\\
         Department of Computer Science and Engineering\\
         PES University\\
         Bengaluru, India\\
         {\tt akhilkred@gmail.com}}
         \parbox{3.3 in}{ \centering Srikanth H R\\
         Department of Computer Science and Engineering\\
         PES University\\
         Bengaluru, India\\
         {\tt srikanthhr@pes.edu}}
}

\begin{document}

\maketitle
\thispagestyle{empty}
\pagestyle{empty}

\begin{abstract}

\textit{Clustering techniques have been the key drivers of data mining, machine learning and pattern recognition for decades. One of the most popular clustering algorithms is DBSCAN due to its high accuracy and noise tolerance. Many superior algorithms such as DBSCAN have input parameters that are hard to estimate. Therefore, finding those parameters is a time consuming process. In this paper, we propose a novel clustering algorithm ‘Bacteria-Farm’, which balances the performance and ease of finding the optimal parameters for clustering. Bacteria-Farm algorithm is inspired by the growth of bacteria in closed experimental farms - their ability to consume food and grow - which closely represents the ideal cluster growth desired in clustering algorithms. In addition, the algorithm features a modular design to allow the creation of versions of the algorithm for specific tasks / distributions of data. In contrast with other clustering algorithms, our algorithm also has a provision to specify the amount of noise to be excluded during clustering.}    \par

\end{abstract}

\textbf{\textit{ Keywords - clustering algorithms;
modular clustering; noise tolerance in clustering; spatial clustering}}


\section{Introduction}

In recent times, clustering has been the center piece of major fields such as data science, machine learning, knowledge discovery, statistics and data mining. In the information age, due to the presence of a plethora of uncleaned, unlabeled data, extraction of insights from this data is very essential in many applications. \par

Clustering is the process of breaking down data into meaningful subdivisions called clusters based on the similarity between data points. The points in a cluster have a higher similarity than the ones across clusters. \par

There is always room for improvement in the clustering paradigm where a newer algorithm is more efficient and effective to a certain distribution of data. One of the most important problems faced while designing a clustering algorithm is choosing the parameters of the algorithm. If the algorithm is susceptible to a tiny change in those parameters, the robustness  of the algorithm is affected. Due to this problem, most of the time is spent on selecting the ideal parameters for the given data than in clustering. Algorithms such as DBSCAN [1] use parameters that are hard to estimate in a short period of time.  \par

\textit{Partitioning clustering algorithms} are the simplest kind of clustering algorithms. The idea is to breakdown the entire data into arbitrary \textit{k} clusters where the partitions optimize a given function. For every cluster, a represent-er in the form of \textit{centroid}, \textit{medoid}, etc. is used to iteratively optimize the clusters with the addition of new data points into the cluster. The advantage of using these algorithms lie in the efficiency of their linearity. But, due to the reliance on the initial configuration of clusters, these algorithms lack robustness. Also, they are not suitable for non-convex data or data with noise. \par

\textit{Hierarchical clustering algorithms} produce a nested structure of clustering data points. They contain two types: \textit{top-down} and \textit{bottom-up}. In \textit{top-down} algorithms, initially, the entire data set is taken as a single cluster and it is sequentially broken down into smaller clusters until they are singleton clusters. On the other hand, \textit{bottom-up clusters} consider every point as a singleton cluster and sequentially combines the data points into bigger clusters than in the previous level. The advantages of using these algorithms lie in the flexibility of choosing the most appropriate number of clusters and their sizes from different levels of clusters. Like the \textit{partitioning cluster algorithms}, they are very sensitive to the presence of noise. Also, they might encounter difficulties in handling convex and large data. Hence, they can prove to be ineffective for real data. \par

\textit{Density based clustering algorithms} group objects / data points based on the density of the locality rather than the proximity between data points. The high density regions are considered as clusters and low density ones as noise. With the advent of density based algorithms, clustering performance was boosted due to the provision of dealing with noise and non-convex data. But, these algorithms are very sensitive to the input parameters as small changes in the values of the parameters can completely shift the structure of clusters. Nevertheless, the performance of density based algorithms is generally greater than partitioning algorithms.\par

\textit{Distribution-based clustering algorithms} group data based on likelihood of data points belonging to a distribution (or cluster). Objects / data points which most likely belong to the same distribution are clustered together. Though their theoretical foundation is sound, they suffer from \textit{over fitting} as complex models are generated easily. Hence, estimation of the complexity of the model is difficult. Moreover, real data may not belong to a precise distribution model and the presence of such models will lead to poor performance in these algorithms [2]. However, distribution-based algorithms work well on complex, spatial data. \par

With a plethora of clustering algorithms with their own advantages and disadvantages, a general algorithm is desired. In this paper, we introduce a modular design to our model to accommodate these various needs of clustering algorithms. In this design, we obtain hyper-parameters for the novel algorithm by pre-clustering a fraction of data with the best standard algorithm for that distribution and fine-tuning these results with our algorithm. Along with this design, the model contains a salient feature to specify the amount of noise to be excluded by the algorithm.  \par

This paper is organized as follows: Related work on clustering, modular algorithms are briefly discussed in Section 2. In Section 3, the design and implementation of the new algorithm are comprehensively explained. In Section 4, the performance evaluation of the algorithm when compared to \textit{k}-means and DBSCAN is presented. Section 5 concludes the paper and some ideas for future research are discussed. 

\section{Related Work and Definitions}

\subsection{Related Work}

One of the first kinds to enter the clustering paradigm are partitioning clustering algorithms. Reference [3] proposes a partitioning algorithm having \textit{k} clusters. Each cluster is represented by a \textit{medoid} and sum of distances within clusters serves as the optimization function. Reference [4] seeks a \textit{local optima} instead of a \textit{global optima} to enhance clustering performance. Reference [5] implements an efficient version of the \textit{Lloyd\textquotesingle s k-means} algorithm to further improve performance.   Reference [6] proposes a \textit{k-d tree} organization of data to efficiently find patterns in the data. Reference [7] proposes a \textit{global k-means} clustering algorithm which incorporates a deterministic global optimization method and employs the \textit{k}-means algorithm as a local search method. The main disadvantage of the work until then was the sensitivity of the algorithm to initial \textit{centroid} positions in the k-means algorithm. By using this method, the issue of randomly selecting initial cluster \textit{centroids} is eliminated and the algorithm proceeds in an incremental way to optimally add a new cluster center to the previous stage. Though this reduces the randomness involved in the k-means algorithm, the sequential addition of a cluster center affects the execution performance. Reference [8] proposes an improved \textit{k}-means algorithm which requires some information on the required domain. With this prerequisite, the algorithm incorporates background information in the form of \textit{instance-level} constraints. Reference [9] proposes a method to reduce the \textit{euclidean} distance calculations in the the original \textit{k}-means algorithm.  \par

Reference [10] proposes a \textit{genetic k-means} algorithm, a hybrid \textit{genetic} algorithm that replaces the \textit{crossover} operation with a \textit{k-means operator} to generate an efficient genetic algorithm for clustering. Reference [11] improvises on the \textit{genetic k-means} algorithm by ensuring the convergence to a global optimum among other  improvisations over its parent version. \par

DBSCAN, a density based algorithm proposed in [1] improved performance drastically with noise handling and design for spatial clustering. It set the benchmark for modern clustering. It introduces a sequential algorithm designed to discover clusters of arbitrary shape. Many versions of DBSCAN were proposed over the years with improvements in efficiency, accuracy and the \textquotesingle power\textquotesingle \ of the algorithm. Reference [12] proposes a sampling-based DBSCAN which improve time efficiency without compromising accuracy. But, there was a problem with this. The sampling cluster might sometimes not represent the population and the clustering deviates from the expected result. Reference [13] presents a hybrid DBSCAN algorithm called l-DBSCAN which uses two \textit{prototypes} to cluster at coarser and finer levels. With this setup, the algorithmic time efficiency and accuracy is greatly improved. Reference [14] introduces ST-DBSCAN which incorporates extensions of DBSCAN to discover clusters for spatial, non-spatial and temporal data as opposed to just spatial data by its parent algorithm. With all the above improvements, the disadvantages of the original DBSCAN were mitigated. Reference [15] uses rough-set theory to create a hybrid clustering technique to derive \textit{prototypes} using the \textit{leader\textquotesingle s clustering} method and use the \textit{prototypes} to derive density based clusters. This split allows a reduction in time complexity from $O(n^2)$ to $O(n)$. Reference [16] introduces MR-DBSCAN which uses the \textit{MapReduce} parallel programming platform to create an efficient implementation of DBSCAN. \par

Reference [17] presents a revised version of DBSCAN that considerably improves DBSCAN’s performance in dense adjacent clusters. Reference [18] presents G-DBSCAN which consists of a GPU accelerated algorithm for density-based clustering. It is evident that the DBSCAN algorithm has evolved since its inception in 1996 but one of its core problems, the presence of parameters which are time-consuming to estimate, is yet to be solved.  \par

Other density based clustering algorithms have also found success, such as the one in [19]. Its algorithm, DBRS, incorporates random sampling and checks a point’s neighborhood to decide whether a point belongs to a cluster or not. Reference [20] presents DBCLASD, a non-parametric algorithm which can form clusters of arbitrary shape by analyzing the distance distributions between data points.

\subsection{Definitions}

\subsubsection{front-runners}

\textit{front-runners} are defined as the points which are ``active'' during the course of the algorithm. They are the points which exist on the periphery of the cluster.

\subsubsection{Dormant points}

Dormant points are points which are not ``active''. All the points in the cluster which are not \textit{front-runners} are considered as dormant points. They are called so because we don't calculate distances to dormant points during the clustering process.

\begin{figure}[hbtp]
    \centering
    \includegraphics[width=0.48\textwidth]{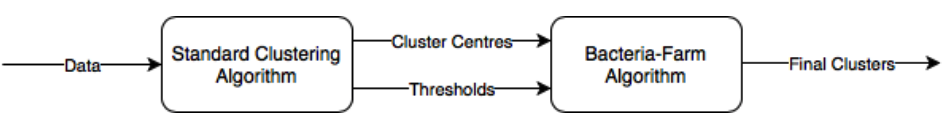}
    \caption{Flow of control in Bacteria-Farm}
    \label{fig:F31}
\end{figure}

\begin{figure*}[tp]
    \begin{center}
        \includegraphics[width=\textwidth]{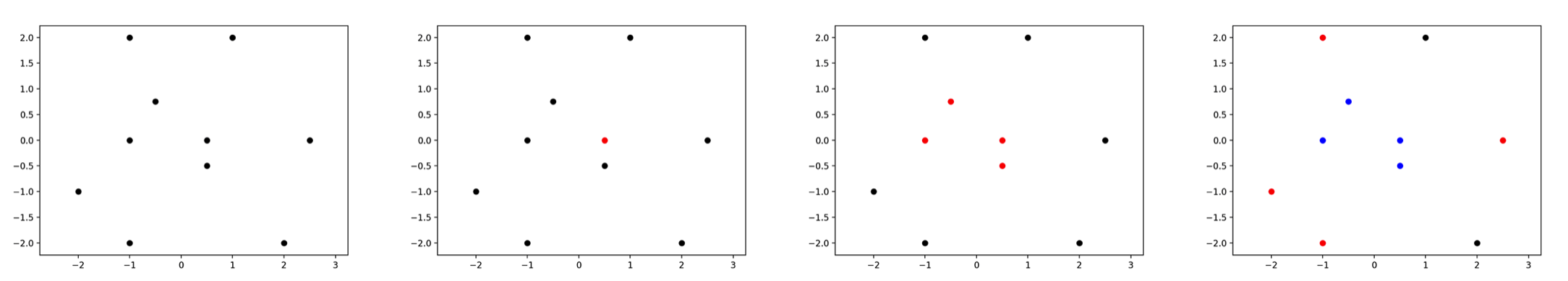}
        \caption{Illustration of the transition of \textit{front-runners} in the Bacteria-Farm algorithm.}
        \label{fig:F32}
    \end{center}
\end{figure*}

\section{A New Modular Spatial Clustering Algorithm}

\subsection{Working of the Algorithm}

\medskip
The model is divided into two phases: \par
\medskip

In the first phase, we sample a portion (typically 20 percent) randomly from the data and apply a standard clustering algorithm on it. This presents flexibility in our model as every distribution of data has its own optimized algorithm which works wonderfully on it. Once the standard algorithm clusters the sample, we extract two parameters from the result - the clustering \textit{centroids} and the proportion of data points in each of the clusters. The proportions act as the threshold for each cluster used in the second phase of the model. \textit{Figure~\ref{fig:F31}} shows the flow of control in our algorithm.\par

In the second phase, the core algorithm is executed. It starts with the \textit{centroid} and expands outward. We have defined a parameter called as \textit{front-runners} which are typically the surface points in a cluster. The distance between every point in the data and the \textit{front-runners} is calculated and the nearest point to the \textit{front-runners} (and hence, to the cluster as they represent the cluster) is selected. This point is included into that cluster. Initially when the number of points in the cluster are less than the number of \textit{front-runners} required, every new point in the cluster becomes a \textit{front-runner}. In the later stages when the number of points in the cluster exceeds the number of \textit{front-runners}, \textbf{the \textit{front-runner} which is closest to the recently selected point goes dormant and is replaced by the new point as the new \textit{front-runner}}. This ensures that the surface points stay as the \textit{front-runners} and the number of \textit{front-runners} stay constant. \textit{Figure~\ref{fig:F32}} illustrates the growth of the cluster in Bacteria-Farm. Iteratively, new points are added to the cluster and the \textit{front-runners} are constantly updated till the exit condition - the number of points in the cluster is equal to the threshold of that cluster (calculated in the first phase) - is satisfied. \par

Once both phases are completed, the clusters are separated from the data and the remaining points - which is noise - are discarded. \par

With the flexibility to specify the number of points a cluster can include in itself, a unique property is observed:  The difference between the total number of points in the data and the sum of the number of points clustered, can be defined as the noise in the data. We use this property for noise specification in the model. When X percent of data is specified as noise to the model, the model excludes X percent of the total data when the proportions of points in each cluster are calculated. \par

\begin{figure}[hbtp]
    \centering
    \includegraphics[width=0.48\textwidth]{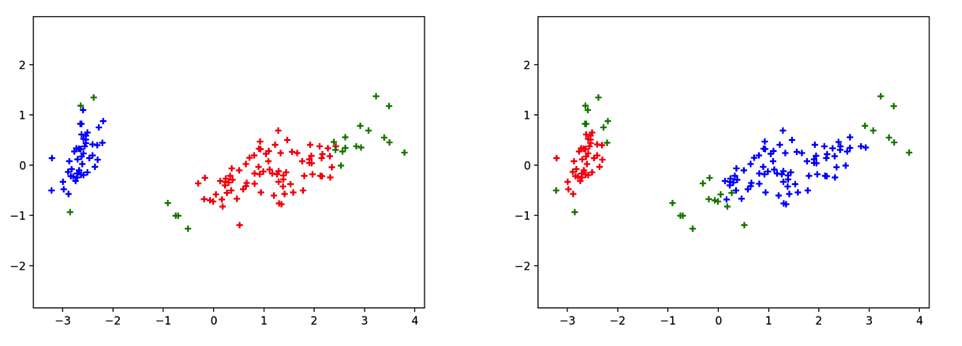}
    \caption{The Iris data set with 15 percent and 20 percent noise specification respectively.}
    \label{fig:F33}
\end{figure}

Suppose there are two clusters with 60 percent and 40 percent as their proportions of number of points. Assume that the noise specified for the model is 10 percent. With the exclusion of noise, the new proportions are 60 percent and 40 percent of the remaining data i.e., 54 percent and 36 percent of the total data. It can be observed that this restriction is on the exit condition of the algorithm and hence, guaranteed (from the design) that the noise specified is excluded only from the periphery of the cluster. \par

\textit{Figure~\ref{fig:F33}} illustrates this property of Bacteria-Farm algorithm. It can be observed that as noise specification increases, the peripheral points are labelled as noise instead of the inner points of a cluster.\par

\subsection{Pseudo-code of the Algorithm}

In the following, we present the pseudo code of the Bacteria-Farm algorithm. Important details are explained separately. The major functions required by Bacteria-Farm are expanded before the core algorithm.  \par

\begin{algorithm}
\caption{sampling}
    \begin{algorithmic}
        \STATE \textbf{Input} data set \textit{df}, number of front-runners \textit{$n_{fr}$}, noise factor \textit{n}
        \STATE \textbf{Output} centroids \textit{centroids}, thresholds \textit{thresholds}
        \STATE
        \STATE \textit{sample} = take a sample of data from the data set \textit{df}.
        \STATE \textit{centroids}, \textit{threshold} = \textit{retrieve-parameters}(\textit{sample}, \textit{n}).
        \STATE \textbf{return} \textit{centroids}, \textit{thresholds}
    \end{algorithmic}
\end{algorithm}

\begin{algorithm}
\caption{retrieve-parameters}
    \begin{algorithmic}
        \STATE \textbf{Input} sample \textit{s}, noise factor \textit{n}
        \STATE \textbf{Output} centroids \textit{centroids}, thresholds \textit{thresholds}
        \STATE
        \STATE Use a standard clustering algorithm to fit the sample data and 	obtain labels for them. The data is divided into \textit{clusters} from this algorithm.
        \FOR{\textit{cluster} in \textit{clusters}}
            \STATE \textit{centroid} = mean of all instances in the \textit{cluster}.
	        \STATE Append \textit{centroid} to the list \textit{centroids}.
	        \STATE \textit{threshold} = ratio of number of points in the \textit{cluster} to the number of points in the \textit{sample} multiplied by the noise factor \textit{n}.
	        \STATE Append \textit{threshold} to the list \textit{thresholds}.
        \ENDFOR
        \STATE \textbf{return} \textit{centroids}, \textit{thresholds}
    \end{algorithmic}
\end{algorithm}

\begin{algorithm}
\caption{Bacteria-Farm}
    \begin{algorithmic}
        \STATE \textbf{Input} data set \textit{df}, number of front-runners \textit{$n_{fr}$}
        \STATE \textbf{Output} clusters \textit{clusters}
        \STATE
        \STATE \textit{centroids}, \textit{thresholds} = \textit{sampling(df,$n_{fr}$)}
        \STATE Initialize a list of lists \textit{clusters} of size \textit{centroids}, to empty lists and each inner list is of size \textit{$n_{fr}$}.
        \FOR{centroid \textit{c} in list \textit{centroids}}
            \STATE \textit{frs} = list of corresponding front-runners for each \textit{c} and initialize the first front-runner as  \textit{c} itself.
            \WHILE{true}
                \IF{size of cluster corresponding to \textit{c} is greater 			than threshold corresponding to \textit{c}}
                    \STATE \textbf{break}
                \ENDIF
                \STATE \textit{minInstance}  = get closest point to the set of front-runners \textit{frs} and add it to current cluster in \textit{clusters}.
                \STATE Replace the closest front-runner \textit{fr} (in \textit{frs}) to the \textit{minInstance}, with \textit{minInstance} itself.
            \ENDWHILE
        \ENDFOR
    \end{algorithmic}
\end{algorithm}

\subsubsection{Additional Explanation}

In retrieve-parameters, we subtract the noise percentage from 100 and multiply that factor with the proportion of points in a cluster obtained from the standard clustering algorithm. This is the threshold for each cluster used in second phase of Bacteria-Farm algorithm. \par

In the core Bacteria-Farm algorithm, we calculate the distance from the front-runners \textit{frs} to every point in the data set \textit{df} and pick the one with the smallest \textit{Euclidean} distance as \textit{minInstance}. The algorithmic complexity of this step is $O(log(n))$ by using spatial indexing. \par

Once the closest point minInstance is chosen, the corresponding front-runner \textit{fr} in \textit{frs} which is closest to \textit{minInstance} is replaced with \textit{minInstance} as the new front-runner and this instance is included to the current cluster. \par

\section{Performance Evaluation}

We evaluate Bacteria-Farm according to the major requirements of clustering algorithms - efficiency, input parameters and ability to cluster data of arbitrary shape. We choose \textit{Silhouette Coefficient} and \textit{Calinski-Harabasz Index} as the performance metrics for evaluation. We compare Bacteria-Farm with established algorithms such as DBSCAN and \textit{k}-Means in terms of efficiency and the fore-mentioned performance metrics. \par

\subsection{Choice of comparison algorithms}

We have selected DBSCAN and \textit{k}-Means for comparison as they are the most popular density-based and partitioning clustering algorithms respectively. We chose DBSCAN as it is an established algorithm for clustering data of arbitrary shape and size. Over time, many versions of DBSCAN have been proposed but the core algorithm remains the same. Hence, we decided to choose the vanilla version of DBSCAN for comparison with the vanilla version of Bacteria-Farm. We have chosen \textit{k}-Means as its time complexity is $O(n)$ and helps in estimating the real performance of the Bacteria-Farm algorithm. \par

\subsection{Choice of performance metrics}

We have selected \textit{Silhouette Coefficient} and \textit{Calinski-Harabasz Index} as the two performance metrics. The definition of these metrics along with the reasons for their selection are given below. \par

\subsubsection{Silhouette Coefficient}

Let $a(i)$ be the average distance between a datum $i$ and all other points in its cluster. $a(i)$ is a measure of the intra-cluster distance. Lower the value of $a(i)$, denser is the cluster and better the assignment of $a(i)$ to the cluster. \par

Let $b(i)$ be the average distance between the datum $i$ and all other points in any other cluster in the data set. $b(i)$ is a measure of the inter-cluster distance. Higher the value of $b(i)$, better the separation of clusters. \par

\textit{Silhouette Coefficient} $s(i)$ can be defined as :

\[s(i) = 
\begin{cases}
    1-a(i) / b(i) & \text{if } a(i) \leq b(i)\\
    0 & \text{if } a(i) = b(i)\\
    a(i) / b(i)-1 & \text{if } a(i) \geq b(i)\\
\end{cases}
\]

For $s(i)$ close to 1, it implies $a(i) \ll b(i)$. A small $a(i)$ means that a datum $i$ is closely matched with other data in the same cluster and a large $b(i)$ indicates that the datum $i$ is poorly matched with data present in other clusters. \par

Therefore, a high value of $s(i)$ can conclude that the data has clustered well. We chose \textit{Silhouette Coefficient} as it has been a good indicator of clustering performance, in the past. \par

\subsubsection{Calinski - Harabasz Index}

Let $SS_B$ be the overall inter-cluster variance, $SS_W$ be the overall intra-cluster variance, $k$ be the number of clusters and $N$ be the number of points in the data set. 

\textit{Calinski-Harabasz Index} $CH_k$ for $k$ clusters (with standard notations) can be defined as :

\begin{figure*}[tp]
    \begin{center}
        \includegraphics[width=\textwidth]{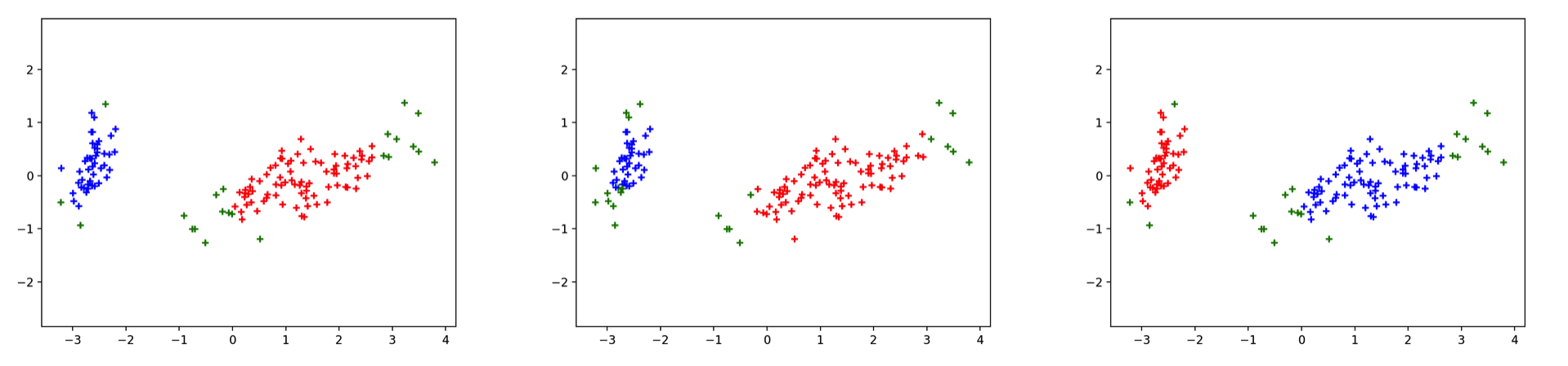}
        \caption{Comparison of results with varying number of \textit{front-runners} (3,5,7) for the Iris data set to demonstrate robustness.}
        \label{fig:F43}
    \end{center}
\end{figure*}

\[CH_k = \dfrac{SS_B}{SS_W} \times \dfrac{N-k}{k-1}
\]

A high value for the first fraction in the above equation indicates that ${SS_B} \gg {SS_W}$. The inference is that the data between clusters are very different from each other and the ones within the same cluster are very similar. This is another indication of good clustering in the data. We chose \textit{Calinski-Harabasz Index} because we can infer that the ratio of variances in the equation is a good indicator of the compact-ability of a cluster. This is a similar to the previous metric.\par

Overall, we have taken two performance indicators (along with time taken for the algorithm to execute) to measure the overall performance of Bacteria-Farm for convex as well as non-convex data. \par

\subsubsection{Input parameters}

It is hard to optimize the parameters for given data in many clustering algorithms. For example, DBSCAN has two parameters  - \textit{Epsilon} distance and \textit{minPts}. To explain those parameters, we define a \textit{core} point. A \textit{core} point is a point which has a minimum number of points within a certain distance from itself. \textit{Epsilon} distance specifies how close the points should be to a core point to consider those points as a part of the cluster and \textit{minPts} specifies how many points should be in the \textit{Epsilon} distance from a point for it to become a core point. Both these parameters are continuous values and are hard to optimize. Users usually resort to running the algorithms multiple times to arrive at the optimal values for these parameters or use optimization techniques to obtain the optimized parameters. This process is time consuming and hence, there is a need for ``better'' parameters. \par

On the other hand, \textit{k}-Means require the number of clusters \textit{apriori} and this is hard to obtain from visual inspection in higher dimension data. \par

After considering these problems, we have devised a different approach to obtain the parameters inherently from our model. As discussed earlier, we have two parts in our model - the first phase which runs a standard algorithm to obtain the clusters and the second phase which runs the core Bacteria-Farm algorithm. Due to the modular design of the model, we can use a parameter-less algorithm in the first phase to obtain clusters. Once the clusters are obtained, the \textit{centroids} of those clusters are sent to the second phase. \par 

Effectively, we have two parameters for Bacteria-Farm : the percent of noise to be specified and the number of \textit{front-runners} desired. Also, the robustness of the model allows for some error in choosing the number of \textit{front-runners}. \textit{Figure~\ref{fig:F43}} illustrates the robustness in the algorithm with varying number of \textit{front-runners}. Many clustering models including DBSCAN fail to account for this error and thus are highly sensitive to small changes in their parameters. \par

\subsubsection{Ability to cluster data of arbitrary shape}

Spatial databases may contain convex, non-convex and other data of arbitrary shape, and good clustering algorithms can cluster any data sufficiently well. We will evaluate DBSCAN and Bacteria-Farm with respect to their ability to cluster data of arbitrary shape. \par

\begin{figure}[hbtp]
    \centering
    \includegraphics[width=0.48\textwidth]{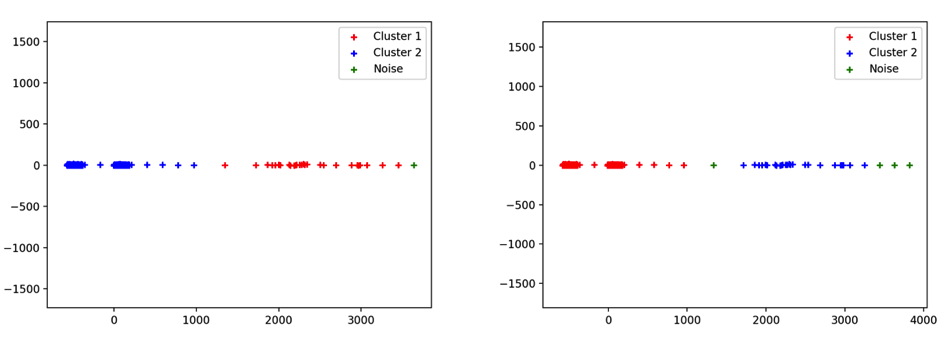}
    \caption{Visual comparison of performance between DBSCAN and Bacteria-Farm for a data set of arbitrary shape.}
    \label{fig:F44}
\end{figure}

We consider a small real data set of small dimensions to illustrate the clustering in Bacteria-Farm algorithm. \textit{Figure~\ref{fig:F44}} shows the clustering comparison between DBSCAN and Bacteria-Farm for the Alcohol data set [21] with 411 instances. \par 

With a noise specification of 1.24 percent for the Bacteria-Farm algorithm, it can be verified visually that the data points are assigned to their correct clusters. It can also be observed that the data points ``seen'' as noise are not included in any of the clusters in the Bacteria-Farm algorithm. Although it can be observed that some points are mis-clustered, they belong to a minority. \par 

\subsubsection{Efficiency}

DBSCAN and Bacteria-Farm are comparable in time complexity of $O(n log(n))$, by using spatial indexing. Whereas, K-Means has a time complexity of $O(n)$. All the measurements are done on a single machine to maintain consistency. \par 

\subsubsection{Performance}

\begin{table}[htbp]
    \begin{center}
        \begin{tabular}{@{}llllll@{}}
        \\
        \hline
        \\
        \textbf{Algorithm}  &\textbf{Silhouette Coefficient}		&\textbf{Calinski-Harabasz Index}				    \\
        \\
        K-Means 		& 0.5842		& 14852.1314		\\
        DBSCAN			& 0.6103   	    & 16657.9614        \\	
        Bacteria-Farm	& 0.6167   	    & 1483.1049			\\	
        \\
        \hline
        \end{tabular}
        \caption{Comparison of algorithms}
        \label{Results}
    \end{center}
\end{table}

We have used 92 different data sets with 200 to approximately 1000 instances to compare the run times and the other performance metrics. \textit{Table~\ref{Results}} tabulates the comparison of performance metric averages between K-Means, DBSCAN and Bacteria-Farm over these data sets. We have chosen data with a small number of instances to verify the inferences and clustering progression in all the three algorithms. It can be derived from the algorithm in the earlier section that the time complexity of Bacteria-Farm is $O(n log(n))$ (by using Spatial Indexing to retrieve distances between points)  and we expect their real run times to be in the same neighborhood. \par 

Since the tasks were not too CPU intensive, the performance comparisons were done on a local computer (Apple MacBook Pro Early 2013) with an Intel HD Graphics 4000 GPU. \par 

The \textit{Silhouette Coefficient} of Bacteria-Farm and DBSCAN are comparable, with Bacteria-Farm performing slightly better. This indicates that the algorithm is able to form clusters with high inter-cluster distance and low intra-cluster distance. Since most linear real datasets such as the one  in \textit{Figure~\ref{fig:F44}} have a low \textit{Silhouette Coefficient}, we have used an average value to compare the overall performance of the algorithms. \par 

On the other hand, a \textit{Calinski - Harabasz} indices of the algorithms are not comparable. This is due to the linear increase of the index with the size of the data. Though it offers some degree of comparison, it is not as effective as \textit{Silhouette Coefficient}. Intuitively, the metrics should have comparable values for data of same shape but of different sizes. And due to this dependency on the variance of size of the data, \textit{Calinski - Harabasz} is used only as a secondary metric. \par

\section{Conclusion and Future Work}

With the modular design, the versatility of the algorithm to cluster the target distribution of data has a significance improvement. With different ``underlying'' algorithms suited for different distributions and types of data, suitable parameters of the data are transferred to the ``core'' Bacteria-Farm algorithm which uses a novel approach to effectively cluster the data. \par 

In this paper, we introduce a novel clustering algorithm / model Bacteria-Farm which is designed to handle noise, introduce parameters which are easy to optimize and display superior performance. Our notion of a cluster depends on the limit of points a cluster can accommodate. The core algorithm is designed to work well with convex and non-convex data. Furthermore, the robustness of the algorithm can be demonstrated by varying the values for \textit{front-runners} as it does not significantly alter the performance of the algorithm. Also, Bacteria-Farm has a provision to specify the amount of noise to exclude from the clusters.  As the clusters extend outward, it is guaranteed that the labelled noise mirror the actual noise in the data. This unique property of noise specification enables applications to generate suitable clusters. \par 

Experiments on real data demonstrate that Bacteria-Farm performs \textit{better}* than algorithms such as DBSCAN and K-Means in the chosen evaluation metrics. The results also indicate Bacteria-Farm performs well for real data. \par 

Future research can include further optimizing the time complexity of retrieval of the closest point to a cluster from $O(log(n))$ and thus, drastically improve the performance of Bacteria-Farm. Also, \textit{projected clustering} can be used to improve Bacteria-Farm for sparse, high dimensional data. \par 

The use of modular design to improve efficiency in other clustering algorithms, and by extension, other paradigms can be explored. Furthermore, we will consider the application of Bacteria-Farm on non-spatial data and explore suitable designs to improve the performance of clustering in non-spatial data. \par

\begingroup
    \fontsize{8pt}{10pt}\selectfont
        * The algorithm performs better than K-Means and is on par, if not better, when compared to DBSCAN; with respect to performance metric Silhouette Coefficient. \par 
\endgroup

\section*{acknowledgement}

We thank the Department of Computer Science and Engineering, PES University for providing the necessary resources to experiment with our models. \par

\end{document}